# The Plant Pathology 2020 challenge dataset to classify foliar disease of apples

Ranjita Thapa[1], Noah Snavely[2], Serge Belongie[2], Awais Khan*[1]

[1]Plant Pathology and Plant-Microbe Biology Section, Cornell University, Geneva, NY, 14456, USA

[2]Cornell Tech, 2 W Loop Road, NY 10044, USA

awais.khan@cornell.edu

**Abstract**

*Apple orchards in the U.S. are under constant threat from a large number of pathogens and insects. Appropriate and timely deployment of disease management depends on early disease detection. Incorrect and delayed diagnosis can result in either excessive or inadequate use of chemicals, with increased production costs, environmental, and health impacts. We have manually captured 3,651 high-quality, real-life symptom images of multiple apple foliar diseases, with variable illumination, angles, surfaces, and noise. A subset, expert-annotated to create a pilot dataset for apple scab, cedar apple rust, and healthy leaves, was made available to the Kaggle community for 'Plant Pathology Challenge'; part of the Fine-Grained Visual Categorization (FGVC) workshop at CVPR 2020 (Computer Vision and Pattern Recognition). We also trained an off-the-shelf convolutional neural network (CNN) on this data for disease classification and achieved 97% accuracy on a held-out test set. This dataset will contribute towards development and deployment of machine learning-based automated plant disease classification algorithms to ultimately realize fast and accurate disease detection. We will continue to add images to the pilot dataset for a larger, more comprehensive expert-annotated dataset for future Kaggle competitions and to explore more advanced methods for disease classification and quantification.*

## 1. Introduction

The U.S. apple industry, annually worth $15 billion, experiences millions of dollars in annual losses due to various biotic and abiotic stresses, ongoing stress management, and multi-year impacts from the loss of fruit-bearing trees. Over the growing season, apple orchards are under constant threat from a large number of insects, fungal, bacterial and viral pathogens, particularly in the Northeastern U.S. **(Figure 1)**. Depending on the incidence and severity of infection by diseases and insects, impacts range from unappealing cosmetic appearance, low marketability and poor quality of fruit, to decreased yield or complete loss of fruit or trees, causing huge economic losses [1]. Early pest and disease detection are critical for appropriate and timely deployment of disease and pest management programs [2]. Disease and pest risk prediction models and management programs are developed based on incidence, severity, and timing of infection, taking into account current and forecasted weather data [3-6]. Incorrect and/or delayed diagnosis and treatment can lead to rapid spread of diseases, and even small instances of insect damage and disease can quickly become a larger and costlier problem, when pathogens multiply rapidly, particularly under favorable environmental conditions. In addition, misdiagnosis can result in either over- or under-use of chemicals, leading to emergence of resistant pathogen strains or increased production costs, environmental and health impacts or otherwise to a significant outbreak. Modern high-density apple orchards, which are usually made up of a few highly susceptible cultivars, are particularly vulnerable to the rapid spread of pathogens, potentially killing the entire orchard [7].

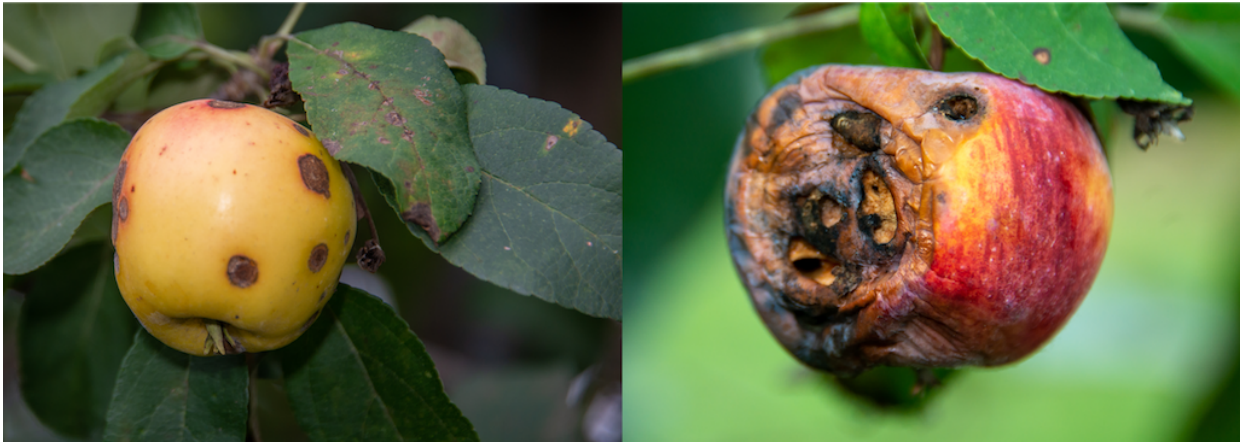

**Figure 1.** Examples of common fungal diseases in an apple orchard impacting cosmetic appearance, fruit quality and yield. Apple scab (*Venturia inaequalis*), cedar apple rust (*Gymnosporangium juniperi-virginianae*) and frogeye leaf spot (*Sphaeropsis malorum*), are very prevalent and economically important fungal diseases of apples across New York State.

Currently, disease and pest detection in commercial apple orchards relies on manual scouting by crop consultants and service providers [8-10]. Unfortunately, very few experienced scouts are available, forcing them to cover many large orchards within a narrow time frame. Scouts require a great deal of expertise and training before they can be efficient and accurate in diagnosing an orchard. Generally, they are first trained using images with symptoms of diseases and insects damage, but due to the presence of a great number of variables in an actual orchard, they need considerable time to familiarize themselves with the many symptom classes caused either due to age and type of infected tissues or stage of the disease pest cycle, as well as changing weather, geographical and cultural differences. Experienced scouts also generally establish a pattern for random sampling to avoid visually evaluating every tree, particularly in large orchards, where areas must be strategically scouted to cover the most important ones. Scouts will look for specific susceptible cultivars such as 'McIntosh', or specific regions of the orchard (*e.g.,* edges etc.). Many symptoms of diseases, pests and abiotic stresses in an apple orchard are distinct enough to differentiate based on visual symptoms alone. However, several disease symptoms look similar enough to each other that it is difficult to accurately determine their cause [11]. At the same time, visual symptoms of a single disease or particular insect can vary greatly between varieties, due to differences in leaf color, morphology, and physiology. Specific temperature, humidity and the physiological developmental stage of a plant also play a crucial role in disease infection and insect development [12, 13]. The shape and form of symptoms can also vary over time as the disease progresses and leaf or fruit tissue ages. In addition to the time spent in the orchard, a scout spends a significant amount of time on each client, entering the scouting report, interpreting results, and providing recommendations for action, such as what kind of spray schedule, spray mixture, and blanket *vs* spot spray, whether pruning is needed, etc. Overall, human scouting is usually time-consuming, expensive and, in some cases, prone to errors.

In recent years, digital imaging and machine learning have shown great potential to speed up plant disease diagnosis [14]. The digital imaging revolution has already created tremendous opportunities in many fields of social and professional life, and much of the world has ready access to a smartphone with an integrated digital camera that can be used to capture high-quality images of disease symptoms. Computer vision methods are being developed to make use of digital images of symptoms for disease classification [15, 16]. These methods combine human expertise and machine learning algorithms to find relationships and visual patterns for grouping and identification. Usually, crowdsourcing platforms are used to collect images with metadata and images are later annotated by experts for training deep neural network models. Once models have been trained, unidentified images can be automatically identified using these models.

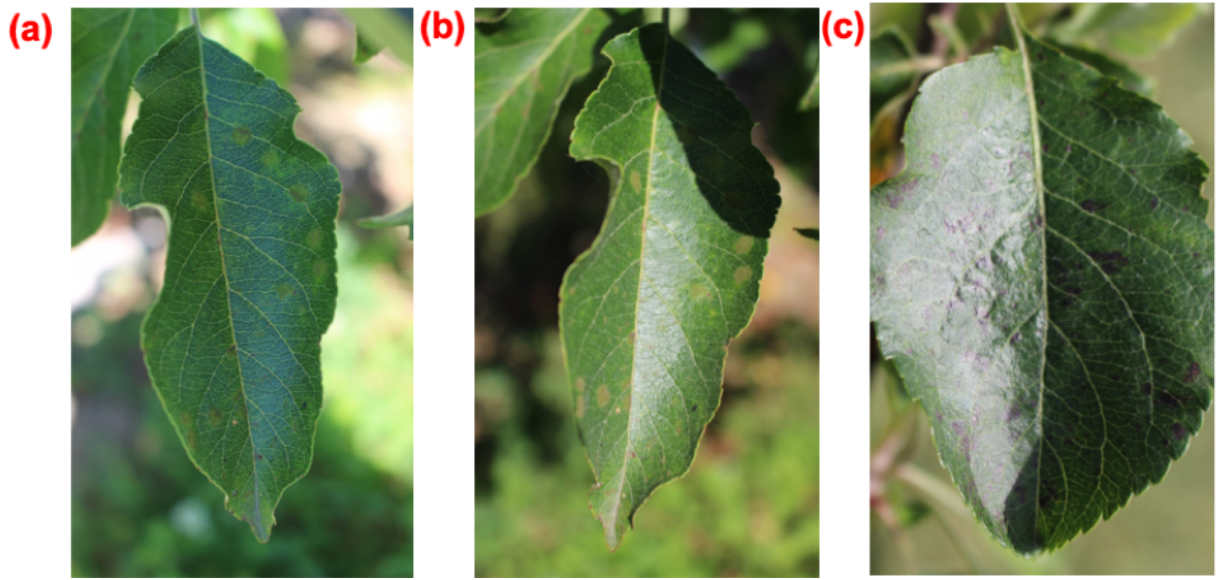

**Figure 2.** Images of disease symptoms on apple leaves captured under different light conditions. a) Indirect sunlight on leaf, b) Direct sunlight on leaf, and c) Strong reflection on leaf.

When using computer vision as a tool for precise disease identification, all potential symptom variables must be accounted for in the digital image database [17]. For example, image capture conditions must include multiple positions and angles of infected tissue in the trees, many shades of lighting, capture sensor types, season and weather results, in addition to each of the different diseases on fruits and leaves of various ages **(Figure 2; Figure 3)**. This makes it difficult, though attractive and theoretically possible, for symptom-based computer vision models to automatically classify disease symptoms with high accuracy without an expert plant pathologist [18]. A large number of high-quality real-life images of disease and insect damage symptoms can be collected and expert-annotated to train computer vision models with high prediction accuracy [17]. Deep convolution neural network (CNN) models and other machine learning models have been previously developed and tested to classify diseased leaf images of crop plants taken in controlled/uniform settings [19-21].

Recent publications [11, 22, 23] have explored the use of computer vision to identify diseases in crops at a more complex level. A CNN based approach was used to detect and distinguish Banana speckle disease from healthy leaves under challenging photographic conditions, including complex backgrounds, different resolutions, different orientations, and various illuminations [15]. A digital image-based portable, interactive, and semi-automatic smartphone application was developed to distinguish diseased tissues from healthy tissues [23]. For assessment of disease severity, these methods require that images are taken in shade either under an umbrella or dense plant canopy to minimize the reflected light from the surface, and preferably with a black background. Cercospora leaf spot (CLS) rater, a computer vision based system, was developed to rate the CLS of sugar beet on a scale of 0-10 using plant-level images taken in the field [22]. An automatic image analysis system was designed to detect and quantify disease symptoms in a variety of leaves with different shapes and sizes. This innovative program was able to successfully quantify the disease severity from images with multiple leaves and save time [11]. However, all of these published methods have limited use in distinguishing a specific disease from a large number of overlapping biotic and abiotic stress symptoms.

The most common foliar diseases of apples are Apple scab, Powdery mildew, Fire blight, Alternaria leaf spot, and Frogeye leaf spot. Apple scab, caused by the fungal pathogen called *'Venturia inaequalis',* is one of the most harmful fungal diseases of apples in temperate regions of world [24]. The typical symptoms of apple scab are visible fungal structures on the leaf and fruit surface **(Figure 1)**. The initial infection appears as black or olive-brown lesions bulging on the upper surface of leaf; later stages show chlorotic sporulating lesions on infected leaves. Apple scab affects not only leaves, but fruits as well, and can cause premature fruit and leaf fall and deformation of fruits. Infected fruits show dark colored, sharply bordered, brown, and corky lesions. Cedar apple rust is caused by a Basidiomycotina fungus called *Gymnosporangium yamadai miyabe.* Early symptoms of the disease are small, light yellow spots on leaves. Later, these spots expand and turn bright orange. The infected leaves get swollen, enlarged, and curled at the edges and in severe cases, premature dropping of leaves occurs. Severe outbreaks of rust pathogen for two to three years can severely injure or kill the trees of susceptible apple varieties [25]. Other foliar diseases of apples, especially Alternaria leaf blotch and frogeye leaf spot, can cause similar-appearing symptoms that can be difficult to distinguish without expert and careful inspection. A CNN model was used to identify four apple leaf diseases (mosaic, rust, brown spot, and Alternaria leaf spot) with an overall accuracy of 97.62% [20]. Dubey et al., (2016) used a K-means clustering method for the detection of infected parts and SVM (Support-Vector Machines) for classification of healthy and infected apple fruits using color, texture, and shape. This multi-class support vector machine has the potential to successfully categorize apple fruit into healthy or infected categories using features extracted from fruits [21].

In this study, we have created an expert-annotated pilot dataset for apple scab, cedar apple rust, and healthy leaves, and used it to train and test a convolutional neural network model for disease classification, and to set-up a Kaggle competition 'Plant Pathology Challenge' ( https://www.kaggle.com/c/plant-pathology-2020-fgvc7 ) as a part of Fine-Grained Visual Categorization (FGVC) workshop at CVPR 2020 (Computer Vision and Pattern Recognition).

## 2. Methods

### 2.1. Apple foliar disease dataset and annotation

We have captured high-quality, real-life RGB images of multiple apple foliar disease symptoms during the 2019 growing season from commercially grown cultivars in an unsprayed apple orchard at Cornell AgriTech, Geneva, NY. Photos were taken using a Canon Rebel T5i DSLR and smartphones under various illumination, angle, surface, and noise conditions **(Figure 2; Figure 3)**. To reflect real-world scenarios, the images were acquired directly from apple orchards during the growing season under various light/angle/surface/noise conditions.

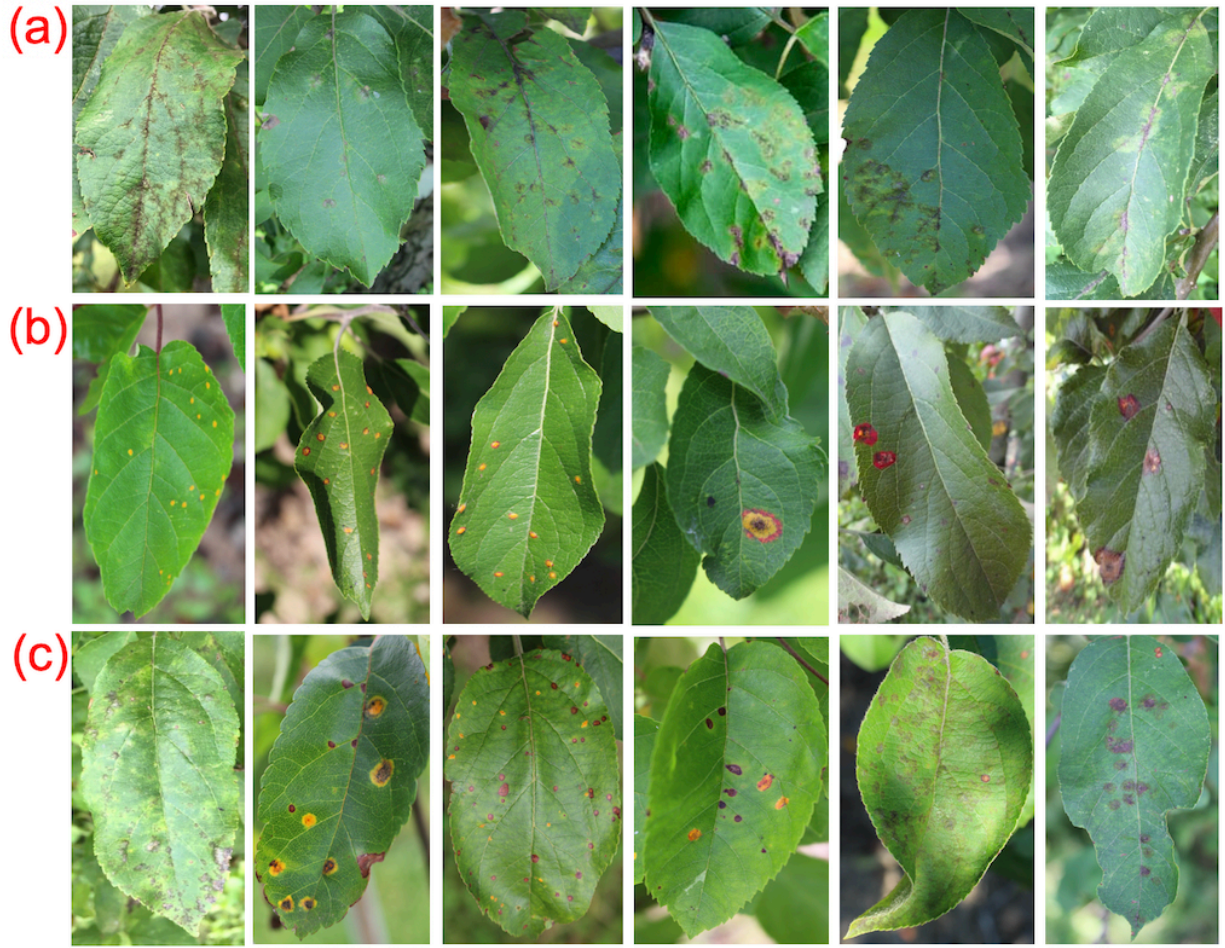

**Figure 3:** Sample images from the dataset showing symptoms of a) Cedar apple rust, b) Apple scab and c) Multiple diseases in a single leaf and effect of differences in lighting, capturing method, age of the leaf and disease symptoms as well as host resistance response.

The complexities of the dataset were increased by including (1) Imbalanced dataset of different disease categories, (2) Non-homogeneous background of images, (3) Images taken at different times of day, (4) Images from different physiological age of the plants, (5) Multiple diseases in the same image, and (6) Different focus of the images. A majority of the pictures taken were of apple scab, cedar apple rust, Alternaria leaf blotch, frogeye leaf spot, and healthy leaves.

We have manually annotated images of apple scab **(Figure 3a)**, cedar apple rust **(Figure 3b)**, and healthy leaves to create annotated disease dataset. The unique and easily distinguishable symptomatic features of apple scab and cedar apple rust were used to annotate them. An expert pathologist confirmed the annotations particularly for images that were difficult to differentiate symptomatically, for example, Alternaria leaf blotch and frogeye leaf spot, and complex disease symptoms from multiple diseases (**Figure 3c**) with similar appearance on the same leaf. The dataset was randomly split into training and stratified test set of 80% and 20%, respectively, such that both datasets have all four disease categories.

## 2.2. Disease classification using a standard convolutional neural network

We have trained an off-the-shelf convolutional neural network on this pilot dataset for classification of apple scab, cedar apple rust, complex disease patterns (leaves with more than one disease in the same leaf), and healthy leaves. Specifically, we took a ResNet50 network pre-trained on ImageNet [26, 27] and fine-tuned the network weights on our annotated disease dataset.

## 2.3. 'Plant Pathology Challenge' for CVPR 2020-FGVC 7 workshop

The current expert-annotated pilot dataset for apple scab, cedar apple rust, complex disease patterns, and healthy leaves was made available to the Kaggle community for 'Plant Pathology Challenge' competition as a part of Fine-Grained Visual Categorization (FGVC) workshop at CVPR 2020 (Computer Vision and Pattern Recognition). In addition to the images, a file with each image_id from the test set was also submitted, to predict a probability for each target variable. The competition was launched on Kaggle and made available to global community to compete on March 9, 2020 and will be open for final submission until May 11, 2020. Submission entries for models will be evaluated on mean AUC (area under the ROC curve) values.

Broader objectives of the 2020 Kaggle competition and future competitions are to train a model using images from the training dataset to 1) Accurately classify a given image from the testing dataset into different disease categories or a healthy leaf; 2) Accurately distinguish between the many diseases, sometimes more than one on a single leaf; 3) Develop an algorithm for quantification of disease severity from the images taken under real-life variable conditions; 4) Deal with rare classes and novel symptoms; 5) Address depth perception—angle, light, shade, physiological age of the leaf; and 6) Incorporate expert knowledge in identification, annotation, quantification, and guiding computer vision to search for relevant features during learning.

## 3. Results

### 3.1. Apple foliar disease dataset and annotation

The pilot dataset consists of 3,651 high-quality annotated RGB images of apple scab **(Figure 3a)** and cedar apple rust symptoms **(Figure 3b)**, complex disease patterns (the leaves with more than one disease in the same leaf), and healthy apple leaves. Images represent real-life field scenarios and were taken under various illumination, angle, surface, and noise conditions **(Figure 2; Figure 3)**. We also have a large number of images of Alternaria leaf blotch and frogeye leaf spot that still need to be annotated. In addition, images taken later in the season and multiple similar looking symptoms on the same leaf created a complex of symptoms that still need expert confirmation and annotation **(Figure 3c)**. The 3,651 RGB images contains 1,200 apple scab, 1,399 cedar apple rust, 187 complex disease, and 865 healthy leaves, respectively. The leaves with complex disease patterns comprised more than one disease in the same leaf. The training dataset had 2,921 images (80%) whereas there were 723 images (20%) images in the test dataset.

### 3.2. Disease classification using a standard convolutional neural network

The overall test accuracy achieved by a ResNet50 network pre-trained on ImageNet is 97% (*i.e.,* 97% of test images are correctly categorized), with the network achieving high accuracy predictions on most categories. The exception is the "complex disease patterns" category (combination of several disease symptoms), which was classified with just 51% accuracy.

### 3.3. 'Plant Pathology Challenge' for CVPR 2020-FGVC 7 workshop

The annotated pilot dataset with 3,651 images of apple scab, cedar apple rust symptoms, complex disease patterns, and healthy leaves was made available to the Kaggle community for 'Plant Pathology Challenge' competition as a part of CVPR 2020 FGVC7 workshop. Since one month of the competition launch, more than 425 teams have participated in the competition and have submitted approximately 1,500 entries. So far, the highest AUC value reported in the public leaderboard is 0.986. There were approximately 41% teams with top entries below 0.95 AUC with the remaining approximately 60% (249) entries are between 0.95 and 0.986 AUC values (**Figure 4**). There are more than 100 entries above 0.97 AUC and more than 15 contending teams reporting an AUC value greater than 0.98.

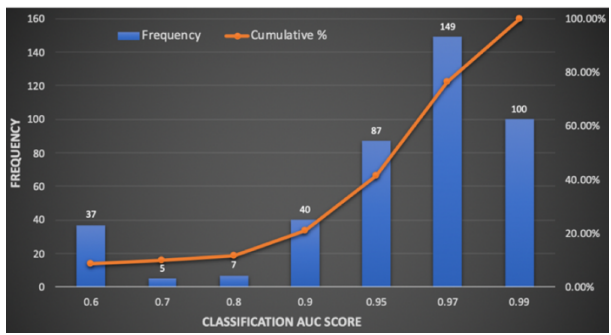

**Figure 4.** Frequency distribution of entries with top AUC (area under the ROC curve) values from each team.

## 4. Discussion

We have developed an expert-annotated disease dataset with 3,651 high quality images of symptoms of apple scab, cedar apple rust, complex disease, and healthy leaves. A curated dataset of disease symptoms and machine learning together offer a great opportunity to develop an active scouting system for more accurate and timely disease management [28]. For example, initiatives like PlantVillage have emerged to collect, host and share large and diverse data sets for use in computer vision to identify diseases [19]. A critical limitation is that the majority of the > 50,0000 images collected by PlantVillage have homogeneous backgrounds [19], which do not accurately reflect conditions and symptoms observed in a real-world scenario. In fact, researchers observed a decrease in accuracy when models trained with images from PlantVillage were applied to predict the accuracy of images collected from other online resources [19, 29]. This suggests that images of single leaves from controlled experiments with uniform light conditions, with few diseases and distinct symptoms do not reflect real-life scenarios in growers' fields.

Early detection of disease symptoms can be used to schedule pesticide application and pruning of infected tissues or eradication of infected apple trees to reduce the chance of severe outbreaks [30, 31]. Current disease identification practices in apple orchards involve scouting by experienced pathologists, shipping of potentially infected samples with visible symptoms to remote laboratories for chemical and PCR tests [32-34], and sharing digital images with experts for identification and recommendation. During a busy growing season, these practices can take so long, from observation, sample collection, shipping, analysis, and identification, that often recommendations are back to growers too late to take necessary control measures. There are also a limited number of extension experts who can provide in-person and on-site identification. Disease diagnosis can be a particular challenge for growers who are new and do not yet have expertise with the crop or for home growers, which can then also be a source of pathogen spread to commercial orchards [35].

Deep learning methods such as deep convolutional neural networks are yielding unprecedented accuracy on visual recognition tasks [36-39]. However, the disease recognition problem is distinct from standard applications of deep learning where there are multiple observations of each object from varying angles. The visual features that define a disease can be small or subtle, localized, and often, symptoms of more than one disease are present on a single leaf [40]. Our pilot dataset consisting of thousands of high-quality, expert annotated images, with both healthy leaves and leaves exhibiting symptoms of apple scab and cedar apple rust, provides an opportunity to test standard deep learning algorithms. A convolutional neural network trained on this pilot leaf disease dataset can achieve 97% test accuracy, giving us confidence in the approach for disease classification. The lower prediction accuracy (51%) of the deep learning method for images with multiple diseases in a single leaf highlights the need to improve methods for image capture and segmentation to increase accuracy of disease classification. We need to (1) handle a broader class of diseases than the three classes in our pilot dataset, (2) generalize to more diverse image conditions than are present in our pilot dataset, (3) achieve even greater than 97% accuracy for a practical system, and (4) visually assess properties beyond presence of disease, such as degree of severity. The performance metrics depicted by the overall accuracy value of our result is relatively lower than the accuracy value reported by Liu et al., (2018) [20]. Although [20] used four foliar diseases of apple *i.e.* Alternaria leaf spot, Mosaic, Rust, and Brown spot, the higher accuracy of 97.62% might be because these diseases have very distinct symptoms and because the images were taken under controlled condition, which do not fully represent the complexities of a natural scenario. The images were taken under controlled conditions, with varying illumination and rotated at different angles to generate multiple pathological images mimicking different light distributions in a natural setting.

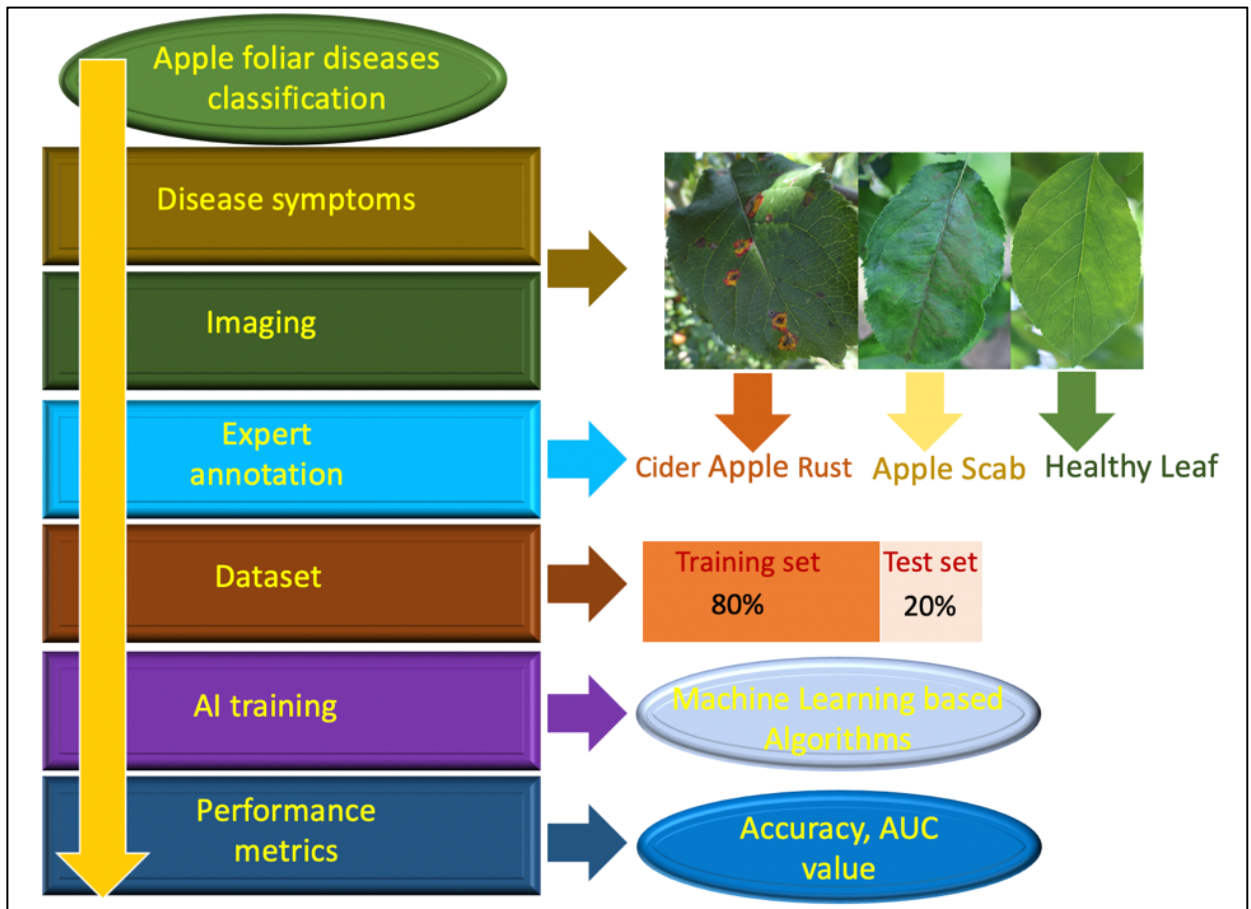

**Figure 5.** A flow diagram showing different steps to develop computer vision models for disease detection based on symptoms.

Currently, our pilot dataset is available through Kaggle for 'Plant Pathology Challenge' competition for FGVC7 workshop CVPR2020, to train models for disease classification **(Figure 5)**. So far, there have been approximately 1,500 entries submitted by 450 teams, with the highest AUC value reported at 0.986 **(Figure 4)**. The large number of submissions with high accuracy demonstrate the promise of this method for automated plant disease diagnosis. A previous plant disease-oriented challenge organized through FGVC6 workshop at CVPR conference was 'iCassava' [41]. The goal of 'iCassava' was to build a robust computer vision algorithm to distinguish between foliar diseases with similar appearances in cassava leaves and to categorize the severity levels of the diseases. A dataset with images of the four diseases: Cassava Brown Streak Disease, Cassava Mosaic Disease, Cassava Bacterial Blight, and Cassava Green Mite was provided to a Kaggle competion. The complexity of the dataset was increased by including 1) images with different backgrounds and scales, 2) images taken at different times of day, 3) some images with poor focus, and 4) images representing mutiple co-occuring diseases on the same plants, as well as by including more unlabeled images (12,595) than labeled images (9,436) [41]. These complexities of the dataset might have contributed to their relatively lower accuracy of 93%. The models trained on this data are likely to have further reduced accuracy when applied to images collected in a farmer's field with a large number of real-world variables.

Going forward, our dataset will continue to contribute towards pushing the state of the art in automatic image classification for identification and quantification of diseases from a large number of symptom classes in various real scenarios **(Figure 5)**. We will continue adding more images captured at a diverse range of angles, lighting, and distances to our pilot dataset to build an even larger, more comprehensive expert-annotated dataset. This will include manually capturing and annotating images of symptoms on apple leaves representing apple scab, fire blight, powdery mildew, cedar apple rust, alternaria leaf blotch, frogeye leaf spot, and marssonina leaf blotch, as well as insects, including apple aphids and mites, on leaves. We will also capture and annotate images of fruit with apple scab, bitter rot, and brown rot.

## 5. Acknowledgements

We acknowledge financial support from Cornell Initiative for Digital Agriculture (CIDA) and special thanks to Zach Guillian and Kai Zhang for help with data collection and testing ResNet model, respectively.